\documentclass[10pt,twocolumn,letterpaper]{article}

\usepackage[pagenumbers]{cvpr}   

\usepackage{algorithm}
\usepackage{algpseudocode}

\definecolor{cvprblue}{rgb}{0.21,0.49,0.74}
\usepackage[pagebackref,breaklinks,colorlinks,allcolors=cvprblue]{hyperref}

\def\paperID{arXiv}
\def\confName{CVPR}
\def\confYear{2026}

\title{GeoComposer: Geometry-Grounded Photographic Composition Instruction}

\author{%
Shuangzhi Li\textsuperscript{1,2,\dag,*} \quad
Qiaoqiao Jia\textsuperscript{1,3,\dag,*} \quad
Xingxin Chen\textsuperscript{1} \quad
Guile Wu\textsuperscript{1} \quad
Dongfeng Bai\textsuperscript{1}\\[2pt]
\textsuperscript{1}Huawei Noah's Ark Lab \quad
\textsuperscript{2}University of Alberta \quad
\textsuperscript{3}University of Waterloo\\[2pt]
{\tt\small shuangzh@ualberta.ca\quad christinajia0803@gmail.com \quad xingxin.chen@huawei.com}\\[1pt]
{\tt\small guile.wu@outlook.com\quad baidongfeng@huawei.com}
}

\begin{document}

\twocolumn[{%
\renewcommand\twocolumn[1][]{#1}%
\maketitle
\begin{center}
    \centering
    \captionsetup{type=figure}
    \includegraphics[width=0.99\textwidth]{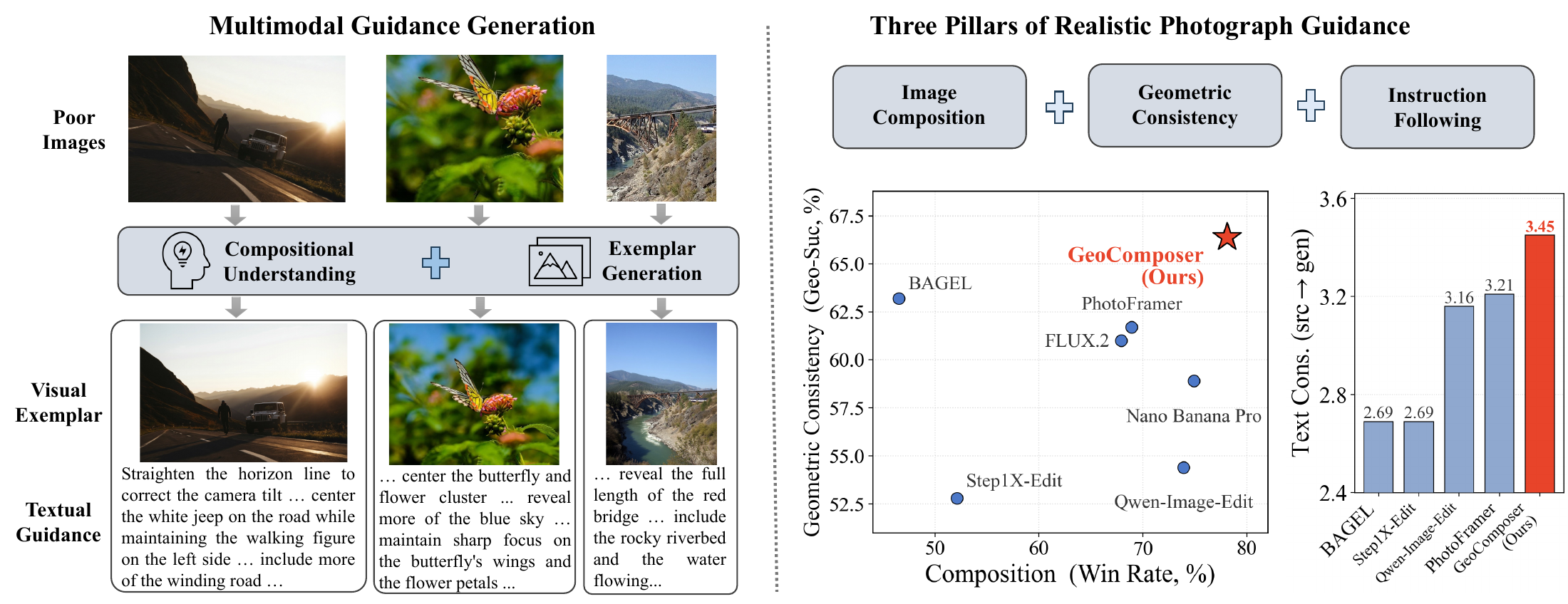}
    \captionof{figure}{Given a poorly composed photograph, GeoComposer generates textual guidance together with a geometry-consistent visual exemplar as the composition improvement guidance. Consistent with textual guidance, the visual exemplar generated by GeoComposer achieves the best balance between composition quality and geometric consistency among existing methods.}
    \label{fig:main}
    \vspace{6pt}
\end{center}%
}]

{\renewcommand{\thefootnote}{\dag}%
\footnotetext{Equal contribution.}}
{\renewcommand{\thefootnote}{*}%
\footnotetext{This work was done during internships at Huawei Canada.}}

\begin{abstract}
Photographic composition aims to provide visual guidance for improving the framing, viewpoint, and spatial arrangement of an image. Early methods primarily rely on image cropping to enhance composition, which is restricted to the viewpoint and spatial arrangement of the input image. Recent methods have explored image understanding and editing to improve composition, but they mainly focus on instruction following and aesthetic quality, overlooking the importance of 3D scene geometry consistency for photographic composition. In this work, we propose GeoComposer, a novel geometry-grounded photographic composition framework that analyzes the composition of a given image to generate textual guidance and synthesizes a visual exemplar that enhances the composition of the given image. To promote geometry-grounded composition, we propose a geometry-aware representation learning mechanism that leverages geometric priors from a visual geometry foundation model to shape the intermediate representations of the composition editing model. This mechanism preserves both global structural relationships and local fine-grained correspondences for geometry-grounded composition. Furthermore, we propose a reinforcement learning strategy guided by a hybrid reward that jointly optimizes instruction following, aesthetic quality, and geometric consistency. This enables the model to generate visual exemplars that faithfully follow the composition instructions while remaining visually appealing and geometrically consistent. Extensive experiments show the superiority of our approach over state-of-the-art methods, highlighting its effectiveness in generating visually appealing and geometrically consistent composition.
Project page: \url{https://geocomposer.github.io/}.
\end{abstract}

\section{Introduction}

Photographic composition plays a fundamental role in visual storytelling.
It involves the careful arrangement of framing, viewpoint, and spatial relationships to enhance the aesthetic quality and narrative impact of an image.
This has motivated the development of intelligent composition assistance methods~\cite{you2026photoframer,farhat2022captain,li2025towards,zhang2018pose} that provide visual guidance, optionally accompanied by textual instructions, to improve photographic composition more intuitively and effectively.
Such guidance is particularly valuable for casual photographers, who may lack expertise in framing and composition.

Early approaches to photographic composition primarily focus on image cropping~\cite{tu2020image,fang2014automatic,hong2021composing}, which select aesthetically pleasing regions from an input image to refine framing and subject placement.
However, these methods are inherently restricted to the viewpoint and spatial arrangement of the input image, allowing only limited post-processing of its framing and spatial layout.
Recent advances in image generation and editing~\cite{wu2025qwen,labs2025flux,xiao2025omnigen,brooks2023instructpix2pix} have enabled more flexible instruction-following image manipulation.
However, most of these methods primarily focus on semantic consistency and instruction fidelity, but remain less effective at following composition-specific instructions and often produce suboptimal compositional results.
In addition, there have been some efforts to develop photographic composition guidance methods that analyze compositional deficiencies of input images and generate aesthetically-enhanced visual exemplars for better composition~\cite{you2026photoframer,li2025towards,du2026aesformer}.
Although these methods have shown promising performance in generating visually appealing composition, they mostly overlook 3D scene geometry consistency in photographic composition, which is crucial for generating viewpoint variations and ensuring spatial coherence.

In this work, we propose GeoComposer, a novel geometry-grounded photographic composition framework that provides both textual guidance and visual exemplars to enhance the composition of input images.
Specifically, GeoComposer employs a composition understanding model to analyze the compositional deficiencies of an input image and generate textual guidance that describes how to improve its composition.
This textual guidance provides users with an interpretable explanation, while also serving as a prompt for a composition editing model to generate a visual exemplar of the suggested adjustment.
However, guided solely by textual instructions, the composition editing model struggles to generate visually appealing and geometrically consistent compositions, as the instructions may not fully capture the underlying 3D scene geometry.
To address this challenge, we propose a geometry-aware representation learning mechanism that shapes the intermediate representation of the composition editing model with geometry-aware tokens from a visual geometry foundation model~\cite{wang2025vggt,wang2026pi} via a global structural supervision and a local correspondence supervision.
This incorporates 3D geometric priors from a visual geometry foundation model into the composition editing model, enhancing its understanding of spatial relationships and viewpoint variations.
Moreover, we introduce a reinforcement learning (RL) strategy that optimizes the composition editing model with a hybrid reward.
In contrast to existing reward formulations that focus primarily on instruction-following and perceptual quality, our hybrid reward explicitly enforces geometric consistency during optimization while also considering instruction fidelity and aesthetic quality.
As a result, our approach generates visual exemplars that are not only faithfully following the composition instructions but also visually appealing and geometrically consistent.
We conduct extensive experiments to evaluate the effectiveness of the proposed GeoComposer framework, demonstrating its superiority over state-of-the-art methods in generating visually appealing and geometrically consistent composition.
Our main \textbf{contributions} are:
\begin{itemize}
    \item We propose a geometry-grounded photographic composition framework that comprehensively resolves the challenges of instruction-following, compositional quality, and geometric consistency in photographic composition.
    \item We introduce a geometry-aware representation learning mechanism that incorporates 3D geometric priors from a visual geometry foundation model into the composition editing model, enhancing its understanding of spatial relationships and viewpoint variations.
    \item We design a reinforcement learning strategy guided by a hybrid reward for optimizing the composition editing model, explicitly enforcing geometric consistency while also considering instruction fidelity and aesthetic quality.
\end{itemize}

\begin{figure*}[t]
    \centering
    \includegraphics[width=0.98\textwidth]{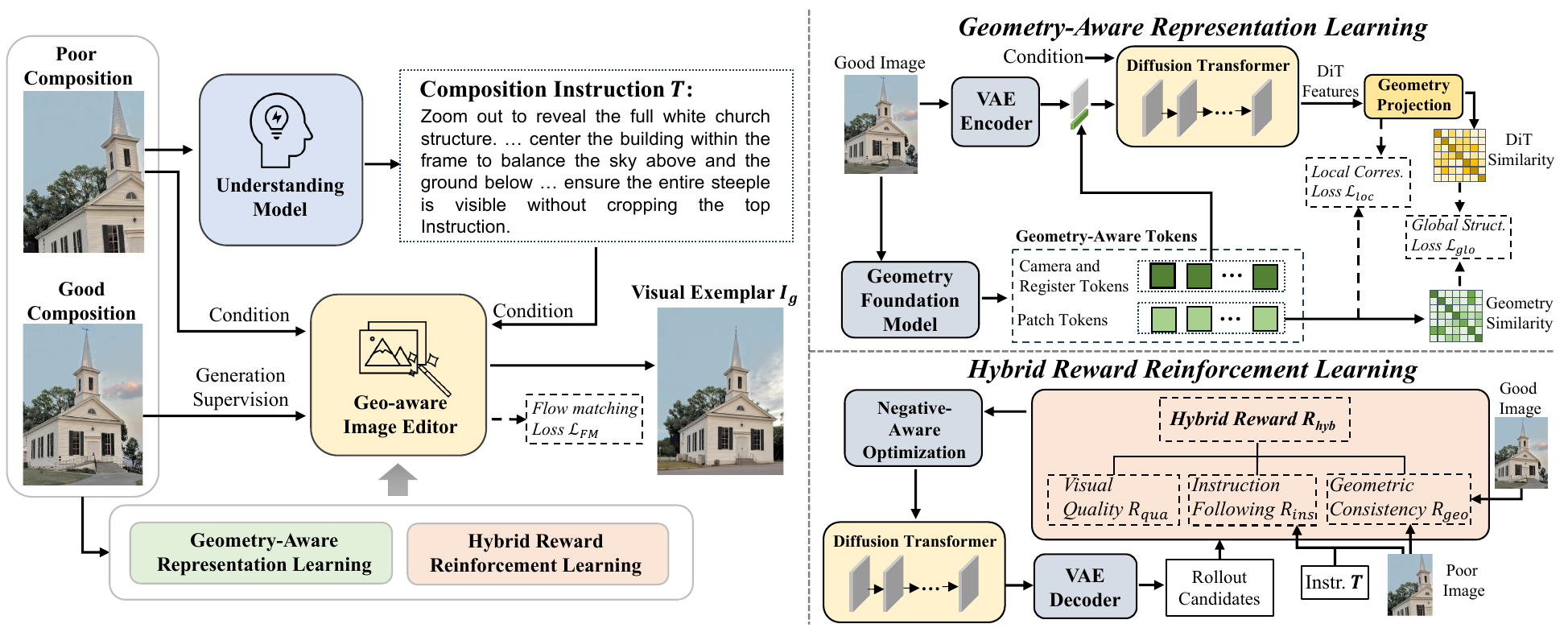}
    \caption{The overall framework of the proposed GeoComposer.}
    \label{fig:framework}
\end{figure*}

\section{Related Work}

\noindent\textbf{Photographic Composition.}
Photographic composition has progressed substantially over the years, evolving from conventional cropping-based methods to generative editing and multimodal composition methods.
Specifically, conventional methods, such as~\cite{tu2020image,fang2014automatic,zeng2020grid,hong2021composing}, approach photographic composition as an image cropping problem.
This paradigm primarily improves composition by selecting aesthetically favorable subregions of the input image through content-aware, subject-aware, or aspect-ratio-aware cropping.
Since these methods do not alter the underlying scene content or viewpoint, they are inherently restricted to the viewpoint and spatial arrangement of the input image, allowing only limited post-processing of its framing and spatial layout.
With rapid advances in generative models, recent methods have moved beyond conventional cropping toward generative image editing and unified multimodal frameworks that jointly support composition understanding and generation~\cite{you2026photoframer,du2026aesformer,yao2026photography}.
Despite these advances, existing methods primarily focus on instruction following and aesthetic quality, while paying limited attention to the geometric consistency of the edited scene.
Consequently, they often struggle to preserve spatial consistency under substantial viewpoint or layout modifications.
Our work addresses the limitations of existing methods by introducing a novel multimodal geometry-grounded photographic composition framework with comprehensive compositional understanding and generation capabilities.
Our framework incorporates geometry-aware representation learning and reinforcement learning guided by a hybrid reward to ensure that the generated visual exemplars are not only aesthetically pleasing and instruction-following but also geometrically consistent with the underlying 3D scene structure.

\noindent\textbf{Image Editing.}
The rapid development of diffusion models has significantly advanced image editing, enabling flexible and high-quality image manipulation.
A typical image-editing paradigm employs a generative model, such as a diffusion model, to transform an input image into a desired output image conditioned on textual prompts or other control conditions~\cite{brooks2023instructpix2pix,labs2025flux,sheynin2024emu,wu2025qwen}.
Existing methods have demonstrated strong instruction-following capabilities while maintaining semantic consistency with the input image.
However, these methods are designed for general-purpose image editing and less effective for photographic composition, as they mainly optimize instruction-following and visual quality rather than compositional quality and geometric consistency.
In contrast, our framework is specifically designed for photographic composition, integrating a composition understanding model to analyze compositional deficiencies and generate composition-aware guidance and a composition editing model to generate visual exemplars that are aesthetically pleasing, instruction-following, and geometrically consistent with the underlying 3D scene structure.

\section{Methodology}

\subsection{The Overall Framework}

Our work focuses on the task of multimodal photographic composition instruction that generates both textual guidance $T$ and visual exemplars $I^g$ to guide users in improving the composition of a given photograph.
An overview of the proposed GeoComposer is depicted in Figure~\ref{fig:framework}.
The GeoComposer framework consists of two main components: \emph{Composition Understanding} and \emph{Visual Exemplar Generation}.

\noindent\textbf{Composition Understanding.}
Given a poorly-composed image $I^p$, we employ a composition understanding model to analyze its compositional deficiencies and generate textual guidance $T$ that describes how the composition can be improved. Rather than producing a vague verdict of image quality, $T$ is required to be \emph{actionable}: it specifies concrete framing adjustments together with brief justifications, so that $T$ can directly condition the downstream editing model.
Since generic Vision-Language Models (VLMs) are not tailored to photographic composition and tend to yield generic or non-executable descriptions, we fine-tune a VLM (e.g., Qwen3-VL~\cite{qwen3vl2025}) on paired data of poorly-composed images and their corresponding improvement instructions.
To make the guidance operation-aware, the model is conditioned on a task prompt $T_{task}$ that indicates the intended type of adjustment (e.g., shifting, zooming, or viewpoint change), and is trained under a mixture of general task prompts so that it can handle both operator-specified intent and fully automatic composition improvement.
Formally, the composition understanding is formulated as 
\begin{equation}
    T = f_{und}(I^p,\, T_{task}),
\end{equation}
where $f_{und}$ denotes the understanding model.
The ground-truth guidance in our training data (please refer to the supplementary material for details) is grounded in real well- and poorly-composed image pairs, and describes the framing transformation that turns the poor composition $I^p$ into its well-composed counterpart. Training on such grounded supervision drives the model to produce concrete, geometry-aware framing-operation guidance that corresponds to a realizable recomposition.
The generated textual guidance serves a dual purpose: providing users with interpretable composition suggestions and prompting the composition editing model to generate a visual exemplar.

\noindent\textbf{Visual Exemplar Generation.}
Next, we employ a diffusion-based composition editing model to generate a visual exemplar $I^g$ that is compositionally improved according to the textual guidance.
\emph{During training}, the composition editing model takes $T$, $I^p$, and $I^t$ as input to learn the compositional transformation from $I^p$ to its corresponding well-composed $I^t$.
Specifically, we first generate multimodal conditions by encoding $T$ and $I^p$ into semantic embeddings using a vision-language encoder~\cite{wu2025qwen} and also encoding $I^p$ into a latent condition using a variational autoencoder (VAE).
In addition, we encode $I^t$ into a latent representation $z_0$ using the same VAE and add Gaussian noise $\epsilon \sim \mathcal{N}(0,I)$ to $z_0$ to obtain the noisy latent $z_t$ at timestep $t$.
Then, we train a Diffusion Transformer (DiT) to denoise $z_t$ under the multimodal conditions, thereby learning the target compositional transformation.
To capture the underlying 3D scene structure, we introduce \emph{Geometry-Aware Representation Learning} to distill geometry-aware knowledge from a pretrained visual geometry foundation model~\cite{wang2025vggt} into the intermediate representations of the DiT through global structural supervision and local correspondence supervision.
Moreover, we introduce \emph{Hybrid Reward-Guided Reinforcement Learning} (RL) to further optimize the DiT with a hybrid reward that jointly evaluates instruction following, visual quality, and geometric consistency.
\emph{During inference}, the model only takes $T$ and $I^p$ as input to generate the composition-enhanced exemplar $I^g$.
\emph{The additional geometry foundation model, reinforcement learning, and the associated computational overhead are only required during training}, while the inference process follows the original diffusion pipeline.

\subsection{Geometry-Aware Representation Learning}

\noindent\textbf{Motivation.}
Since the textual guidance generated by the composition understanding model may not fully capture the underlying 3D scene geometry, the composition editing model struggles to generate geometrically consistent compositions when guided solely by textual instructions.
We therefore propose to distill geometry-aware knowledge from a visual geometry foundation model~\cite{wang2025vggt,wang2026pi} into the intermediate representations of the DiT.
Unlike existing representation alignment methods~\cite{yu2024representation,singh2025matters,xu2026beyond} that primarily focus on transferring semantic knowledge from vision foundation models, our approach aims to distill geometric knowledge from geometry foundation models to enhance geometry awareness for photographic composition through two complementary supervisions:
1) a \emph{global structural supervision} that enforces the DiT to maintain the same global structural information as the geometry-aware representations;
2) a \emph{local correspondence supervision} that encourages the DiT features to align with the geometry-aware representations at a local level.

\noindent\textbf{Global Structural Supervision.}
Given a pair of images $(I^p, I^t)$, we employ a pre-trained visual geometry foundation model to extract geometry-aware representations $F_{\mathrm{geo}}$ for the target image $I^t$.
$F_{\mathrm{geo}}$ consists of geometry-aware patch tokens $F^{\mathrm{patch}}$, a camera token $F^{\mathrm{cam}}$, and register tokens $F^{\mathrm{reg}}$.
Meanwhile, we feed the noisy latent $z_t$ of $I^t$ into the DiT to obtain intermediate features $F^{\mathrm{DiT}}$.
Then, we employ a projection head $P(\cdot)$ to project $F^{\mathrm{DiT}}$ into the latent space of the geometry foundation model, resulting in the projected DiT features $\hat{F}^{\mathrm{DiT}}$.
Instead of directly aligning the features using L2 loss like~\cite{yu2024representation}, we calculate the relational similarity among $F^{\mathrm{patch}}$ to capture the global structural information of the scene and enforce $\hat{F}^{\mathrm{DiT}}$ to maintain the same global structural information.
Specifically, we compute pairwise cosine similarity matrices $S_{\mathrm{geo}}$ and $S_{\mathrm{DiT}}$ over the normalized features as follows:
\begin{equation}
S_{\mathrm{geo}}=F^{\mathrm{patch}}(F^{\mathrm{patch}})^{\top}, \quad S_{\mathrm{DiT}}=\hat{F}^{\mathrm{DiT}}(\hat{F}^{\mathrm{DiT}})^{\top},
\end{equation}
where $F^{\mathrm{patch}}$ and $\hat{F}^{\mathrm{DiT}}$ are normalized before computing $S_{\mathrm{geo}}$ and $S_{\mathrm{DiT}}$.
We minimize the mean squared error between $S_{\mathrm{geo}}$ and $S_{\mathrm{DiT}}$ to enforce global structural consistency as:
\begin{equation}
\mathcal{L}_{\mathrm{glo}}=\frac{1}{N^2}\left\|S_{\mathrm{DiT}}-S_{\mathrm{geo}}\right\|^2,
\end{equation} 
where $N$ is the number of patch tokens.
In addition, since $F^{\mathrm{cam}}$ and $F^{\mathrm{reg}}$ provided by the geometry foundation model also encode global scene-level geometric priors, we can further incorporate them into the DiT features to enhance the global structural supervision.
To achieve this, we concatenate $F^{\mathrm{cam}}$ and $F^{\mathrm{reg}}$ with $z_0$ before adding noise, and feed the concatenated features into the DiT.
These special tokens provide additional global geometric context to the DiT, enabling it to better capture the underlying 3D scene structure and improve its geometry awareness.
During inference, we use Gaussian noise to replace these special tokens, allowing the DiT to generate compositionally improved images without requiring a geometry foundation model.

\noindent\textbf{Local Correspondence Supervision.}
Although the global structural supervision facilitates global geometric consistency, it does not explicitly enforce local feature correspondence between the DiT features and the geometry-aware representations.
We thus introduce local correspondence supervision to encourage the DiT features to align with the geometry-aware representations at a local level.
Specifically, we measure the cosine similarity $S_{\mathrm{cos}}$ between $\hat{F}^{\mathrm{DiT}}$ and $F^{\mathrm{patch}}$, and calculate the local correspondence loss $\mathcal{L}_{\mathrm{loc}}$ as:
\begin{equation}
\mathcal{L}_{\mathrm{loc}}=1-S_{\mathrm{cos}}(\hat{F}^{\mathrm{DiT}}, F^{\mathrm{patch}}).
\end{equation}

\noindent\textbf{Overall Training Objective.}
For the DiT-based composition editing model, we employ the flow matching objective $\mathcal{L}_{\mathrm{FM}}$~\cite{lipman2022flow} and further augment it with the proposed geometry-aware supervision $\mathcal{L}_{\mathrm{glo}}$ and $\mathcal{L}_{\mathrm{loc}}$.
The overall training objective is defined as:
\begin{equation}
\mathcal{L}_{\mathrm{geo}}=\mathcal{L}_{\mathrm{FM}}+\lambda_{\mathrm{glo}}\mathcal{L}_{\mathrm{glo}}+\lambda_{\mathrm{loc}}\mathcal{L}_{\mathrm{loc}},
\end{equation}
where the weighting coefficients $\lambda_{\mathrm{glo}}$ and $\lambda_{\mathrm{loc}}$ balance the contributions of the global and local geometry supervision relative to the flow matching objective.

\subsection{Hybrid Reward-Guided Reinforcement Learning}

\noindent\textbf{Motivation.}
Although \emph{Geometry-Aware Representation Learning} distills 3D geometric priors into the DiT, its diffusion objective remains a denoising trajectory that regresses toward the ground-truth reference.
However, photographic composition is inherently multi-solution. That is, for a poorly-composed image, there may exist many potential reframings.
Thus, supervising the model to reproduce one particular target both suppresses this diversity and caps the attainable quality at that of the reference.
Moreover, the training objective $\mathcal{L}_{\mathrm{geo}}$ cannot jointly enforce instruction-following, aesthetic quality, and geometric consistency, as it only provides a reconstruction-based supervision that encourages the model to reproduce the reference image.
We therefore further optimize the composition editing model through \emph{Hybrid Reward-Guided Reinforcement Learning}, which replaces single-target regression with reward objectives evaluated directly on generated images.
This lets the RL policy explore the space of valid compositions \emph{beyond ground truth}, producing edits that are simultaneously instruction-aligned, visually realistic, and geometrically coherent.

\noindent\textbf{Rollout Generation.}
Given a poorly-composed image and the corresponding editing instruction, we first generate $K$ composition candidates $\mathcal{Y}=\left\{I^c_1,I^c_2,\cdots,I^c_K\right\}$ for each input by performing rollout sampling from the current diffusion policy~\cite{zheng2025diffusionnft}.
Compared to using a single generated sample, the rollout strategy enables the model to explore diverse composition solutions and provides richer supervision for policy optimization.

\noindent\textbf{Hybrid Reward.}
Existing RL approaches for image editing typically rely on an editing reward that evaluates instruction following and semantic quality.
However, for photographic composition, a comprehensive evaluation should consider not only instruction fidelity, but also aesthetic quality and geometric consistency.
In light of this, we introduce a hybrid reward $R_{\mathrm{hyb}}$ that encourages the diffusion policy to generate edits that are instruction-following, visually appealing, and geometrically consistent:
\begin{equation}
R_{\mathrm{hyb}}=\lambda_{\mathrm{ins}}R_{\mathrm{ins}}+\lambda_{\mathrm{qua}}R_{\mathrm{qua}}+\lambda_{\mathrm{geo}}R_{\mathrm{geo}},
\end{equation}
where $R_{\mathrm{ins}}$, $R_{\mathrm{qua}}$, and $R_{\mathrm{geo}}$ denote the instruction-following, aesthetic-quality, and geometry-consistency rewards, respectively, while $\lambda_{\mathrm{ins}}$, $\lambda_{\mathrm{qua}}$, and $\lambda_{\mathrm{geo}}$ are weighting coefficients that balance their respective contributions.

Specifically, for $R_{\mathrm{ins}}$, we use a VLM to jointly process the input image, the generated image, and the textual instruction to estimate the degree of editing success while penalizing unnecessary modifications to the original content:
\begin{equation}
    R_{\mathrm{ins}} = f_{\mathrm{vlm}}(I^p, I^c, T),
\end{equation}
where $f_{\mathrm{vlm}}$ is a VLM that outputs an instruction-following score.
For $R_{\mathrm{qua}}$, we employ a VLM to evaluate the generated image in terms of naturalness and aesthetic quality as:
\begin{equation}
    R_{\mathrm{qua}} = f_{\mathrm{vlm}}(I^c, T^p),
\end{equation}
where $f_{\mathrm{vlm}}$ is a VLM that outputs a quality score, and $T^p$ is a prompt that instructs the VLM to assess the quality of the generated image.
Although these two rewards encourage semantically correct and visually appealing outputs, they do not enforce geometric plausibility.
Consequently, the results may favor improvements that increase the editing score at the expense of 3D scene consistency.
To resolve this problem, we introduce a geometry-aware scene-consistency reward $R_{\mathrm{geo}}$ derived from a visual geometry foundation model.
We use a geometry foundation model to extract patch representations together with intermediate attention features.
Then, we compute patch-level geometric relevance using two complementary cues~\cite{han2025emergent}, including (1) attention relevance derived from global attention patterns of the model and (2) geometry-aware feature similarity between patch representations.
The resulting patch-level relevance values are aggregated using generalized mean (GeM) pooling~\cite{radenovic2018fine} to obtain image-level attention and feature relevance scores.
Specifically, let $S^{x{\rightarrow}y}$ denote the directed geometry relevance score from image $I^x$ to image $I^y$.
Each image $I^x$ has its patch tokens $F_x$ along with query features $Q_x$ and key features $K_x$ probed from the attention layer.
The attention relevance sums, for each source patch, the softmax attention mass it assigns to patches $F_y$:
\begin{equation}
S_{\mathrm{attn}}^{x\rightarrow y}=\mathrm{GeM}_{p}
\left(
\sum_{i\in{F}_y}
\mathrm{softmax}
\left(
\frac{Q_x^{\top} K_y}{\sqrt{d}}
\right)
\right),
\end{equation}
where $d$ is the head dimension and the softmax is normalized over all tokens.
The feature similarity takes, for each source patch, its maximum cosine match in $F_y$:
\begin{equation}
S_{\mathrm{feat}}^{x\rightarrow y}
=
\mathrm{GeM}_{p}
\left(
\phi\left(\frac{F_x^{\top} F_y}{\|F_x\|\|F_y\|} \right)
\right),
\end{equation}
where $\phi(\cdot)$ denotes the log-sum-exp for obtaining a soft maximum. The two complementary measures are combined as:
\begin{equation}
S^{x\rightarrow y}
=
\lambda_{\mathrm{attn}}
S_{\mathrm{attn}}^{x\rightarrow y}
+
\lambda_{\mathrm{feat}}
S_{\mathrm{feat}}^{x\rightarrow y},
\end{equation} 
where $\lambda_{\mathrm{attn}}$ and $\lambda_{\mathrm{feat}}$ control relative contributions of two measures.
To obtain a symmetric geometry consistency measure, we average the bidirectional relevance scores between the generated image and both the source and target images,
\begin{equation}
R_{\mathrm{src}}
=
\frac{
S^{p\rightarrow c}
+
S^{c\rightarrow p}
}{2},
\qquad
R_{\mathrm{tgt}}
=
\frac{
S^{t\rightarrow c}
+
S^{c\rightarrow t}
}{2},
\end{equation}
where $S^{p\rightarrow c}$ and $S^{c\rightarrow p}$ denote the directed relevance scores between the source and generated images, while $S^{t\rightarrow c}$ and $S^{c\rightarrow t}$ denote the directed relevance scores between the target and generated images.
$R_{\mathrm{src}}$ encourages the generated image to preserve the geometric structure of the source image, whereas $R_{\mathrm{tgt}}$ encourages geometric alignment with the desired composition represented by the target image.
Finally, the geometry-aware reward is computed as:
\begin{equation}
R_{\mathrm{geo}}
=
w_sR_{\mathrm{src}}
+
w_tR_{\mathrm{tgt}},
\end{equation} 
where $w_s$ and $w_t$ control the relative importance of source preservation and target alignment, respectively.
The geometry reward penalizes outputs that achieve high editing scores through spatially inconsistent changes, solving reward hacking and encouraging generated images that are instruction-aligned, visually realistic, and geometrically coherent.

\begin{table*}[!t]
\centering
\small
\setlength{\tabcolsep}{3.5pt}
\begin{tabular}{l cc cc cc cc c}
\toprule
& \multicolumn{2}{c}{Composition} & \multicolumn{2}{c}{Image Quality} & \multicolumn{2}{c}{Geo. Consistency} & \multicolumn{2}{c}{Text Guidance} & Human \\
\cmidrule(lr){2-3} \cmidrule(lr){4-5} \cmidrule(lr){6-7} \cmidrule(lr){8-9} \cmidrule(lr){10-10}
Method & WinRate (\%)$\uparrow$ & PF-ass$\uparrow$ & QAlign$\uparrow$ & DeQA$\uparrow$ & Geo-Suc (\%)$\uparrow$ & MET3R$\downarrow$ & src$\rightarrow$gen$\uparrow$ & src$\rightarrow$GT$\uparrow$ & Overall$\uparrow$ \\
\midrule
BAGEL & 46.5 & 2.588 & 2.823 & 3.598 & \underline{63.2} & 0.246 & 2.69 & 2.57 & 43.3 \\
Step1X-Edit & 52.1 & 2.704 & 2.870 & 3.496 & 52.8 & 0.231 & 2.69 & 2.30 & -- \\
Qwen-Image-Edit & 73.9 & 2.987 & 3.272 & 3.869 & 54.4 & 0.257 & 3.16 & 2.60 & 59.4 \\
FLUX.2 & 67.9 & 2.948 & 3.133 & 3.820 & 61.0 & 0.233 & -- & -- & 40.6 \\
PhotoFramer & 68.9 & 2.957 & 3.043 & 3.812 & 61.7 & 0.231 & \underline{3.21} & \underline{2.69} & 63.3 \\
Nano Banana Pro & \underline{74.9} & \underline{3.081} & \underline{3.323} & \textbf{4.041} & 58.9 & \underline{0.191} & -- & -- & \underline{63.9} \\
\midrule
GeoComposer (Ours) & \textbf{78.1} & \textbf{3.133} & \textbf{3.381} & \underline{4.012} & \textbf{66.4} & \textbf{0.184} & \textbf{3.45} & \textbf{3.28} & \textbf{85.6} \\
\bottomrule
\end{tabular}
\caption{Quantitative comparison with state-of-the-art methods on the constructed validation set.
The best and second-best results are marked in \textbf{bold} and \underline{underline}, respectively.
``--'' denotes unavailable results, i.e., pure image generators emit no reasoning text for text-guidance evaluation, and Step1X-Edit has no human-study result.}
\label{tab:main}
\end{table*}

\begin{figure*}[t]
\centering
\includegraphics[width=0.99\textwidth]{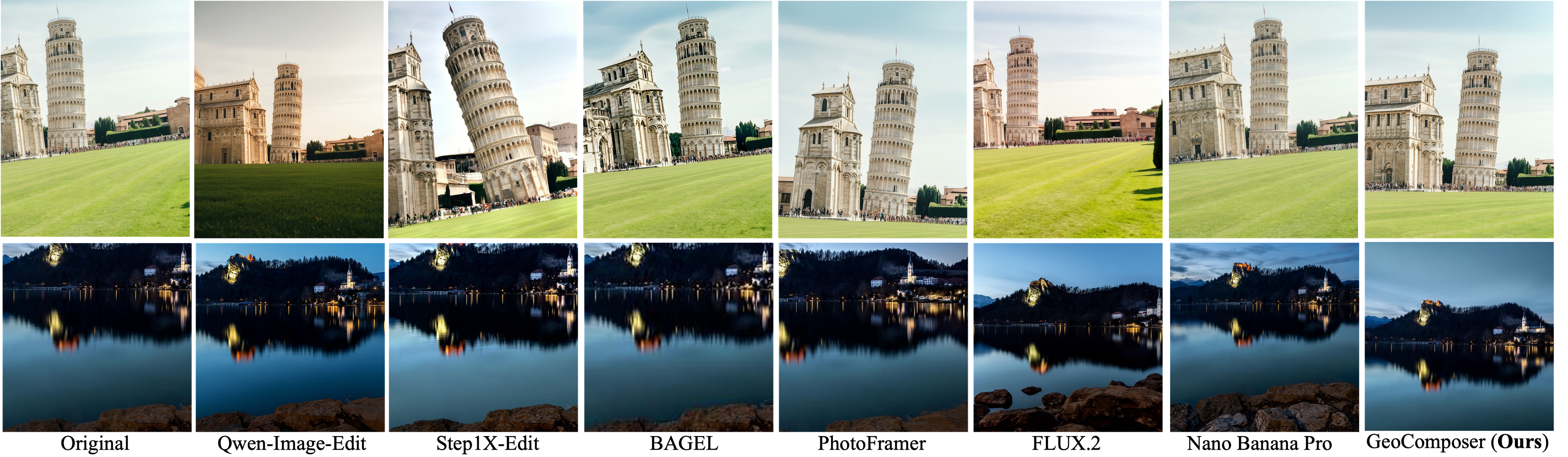}
\caption{Qualitative comparison with state-of-the-art methods.}
\label{fig:qualitative}
\end{figure*}

\noindent\textbf{Negative-Aware Optimization.}
Following DiffusionNFT~\cite{zheng2025diffusionnft}, we compute a group-relative advantage for each candidate by normalizing its hybrid reward against the statistics of the $K$ rollouts generated for the same input.
The normalized advantage weights a continuous interpolation between a positive and a negative velocity target for each candidate: candidates scoring above the group average are drawn toward the positive target, while those scoring below it are pushed toward a negative target constructed by mirroring the current velocity prediction about that of the reference policy.
Within the flow matching framework, these mirrored targets provide counterfactual supervision that penalizes low-reward candidates, whether geometrically inconsistent or otherwise inferior, guiding the diffusion policy toward instruction-aligned, visually realistic, and geometrically consistent image generation.

\section{Experiments}

\subsection{Experimental Setup}

\noindent\textbf{Datasets.}
Following the data construction in PhotoFramer~\cite{you2026photoframer}, we build the dataset on top of public datasets, i.e., CPC~\cite{wei2018good}, GAIC~\cite{zeng2020grid}, FLMS~\cite{fang2014automatic}, CUHK-ICD~\cite{yan2013learning}, and Unsplash~\cite{unsplash}, from which composition pairs and their text guidance are derived.
We train and validate GeoComposer on the constructed dataset of 34,129 $\langle$poor, good, text-guidance$\rangle$ triplets, with the full construction pipeline detailed in the supplementary material.
The dataset is split at the source-image level into 33,229 training and 900 validation triplets.
To further assess generalization beyond the constructed dataset, we additionally evaluate on a held-out benchmark, DL3DV~\cite{ling2024dl3dv}, which is disjoint from the training data.

\noindent\textbf{Evaluation Metrics.}
We evaluate along four automatic axes together with a human study.\footnote{Detailed metric protocols, per-baseline descriptions, and full implementation details are provided in the supplementary material.}
For \emph{composition}, Comp.~WinRate is the pairwise win rate of the edited image against the source, and PF-ass is the single-image PhotoFramer composition score~\cite{you2026photoframer}.
For \emph{image quality}, QAlign~\cite{wu2024qalign} and DeQA~\cite{you2025deqa} measure no-reference aesthetic and general image quality.
For \emph{geometric consistency}, MET3R~\cite{asim2025met3r} measures multi-view 3D consistency (lower is better), and Geo-Suc, an LLM-judge metric, measures whether the edit realizes the correct viewpoint change rather than merely altering the image.
For \emph{text-guidance accuracy}, Text Cons.~(src$\rightarrow$gen) and (src$\rightarrow$GT) score how faithfully the reasoning text describes the method's own edit and the human reframing~\cite{you2026photoframer}.
The human study reports the fraction of edits judged both geometrically valid and better composed than the source.
Unless noted, all LLM-judge metrics use GPT-5.4 as the judge.

\noindent\textbf{Baselines.}
We compare with the general-purpose editors BAGEL~\cite{deng2025emerging}, Step1X-Edit~\cite{liu2025step1x}, Qwen-Image-Edit~\cite{wu2025qwen}, and FLUX.2~\cite{labs2025flux2}, the proprietary Nano Banana Pro (Gemini 3 Pro Image)~\cite{google2025nanobananapro}, and the composition-specific PhotoFramer~\cite{you2026photoframer}.

\subsection{Comparison with State-of-the-Art Methods}

Table~\ref{tab:main} reports the quantitative comparison on the constructed validation set.
GeoComposer achieves the best performance on seven of the eight metrics and remains highly competitive on the remaining one.
For composition, our method attains a win rate of 78.1\%, together with the best PF-ass score.
For image quality, our method achieves the best QAlign score and a DeQA score on par with the best Nano Banana Pro.
For geometric consistency, our method achieves the best MET3R and the best Geo-Suc, indicating that GeoComposer performs the intended photographic changes while preserving the underlying 3D scene structure.
Notably, competitive baselines exhibit a clear trade-off between the two geometry metrics: BAGEL attains a relatively high Geo-Suc but a near-50\% win rate, suggesting conservative edits, whereas Nano Banana Pro edits more aggressively at the cost of geometric correctness.
In contrast, GeoComposer is the only method that ranks first on both composition and geometric consistency, demonstrating that our geometry-grounded design improves composition without sacrificing spatial coherence.
For text-guidance accuracy, our method produces the most faithful and correct textual guidance, outperforming PhotoFramer on both Text Cons.~(src$\rightarrow$gen) and Text Cons.~(src$\rightarrow$GT).
The human study further corroborates these results, where GeoComposer is judged an overall improvement over the source in 85.6\% of cases, far exceeding the strongest baseline at 63.9\%.
As shown in Fig.~\ref{fig:qualitative}, GeoComposer also produces the most visually appealing and geometrically consistent composition among all methods.
GeoComposer further generalizes to the held-out DL3DV benchmark, achieving the best geometric consistency, as detailed in the supplementary material.

\subsection{Ablation Studies}

\noindent\textbf{Effectiveness of Each Component.}
Table~\ref{tab:ablation_component} presents the ablation of the three key components of GeoComposer, i.e., the composition understanding model, geometry-aware representation learning (Stage I), and hybrid reward-guided reinforcement learning (Stage II), on top of the Qwen-Image-Edit backbone.
Introducing the composition understanding model improves composition and image quality, as the generated guidance drives more substantial compositional edits, but this comes at the cost of degraded geometric consistency.
Adding geometry-aware representation learning substantially restores geometric consistency while maintaining comparable quality, validating the effectiveness of distilling 3D geometric priors into the DiT.
Hybrid reward-guided reinforcement learning further improves all four metrics simultaneously, confirming that image-level reward optimization complements representation-level geometry distillation.

\noindent\textbf{Composition Understanding Model.}
Table~\ref{tab:ablation_understand} compares different composition understanding models for generating textual guidance, with the composition editing model kept identical.
Our fine-tuned understanding model outperforms both the zero-shot Qwen3-VL and the guidance produced by PhotoFramer, demonstrating that composition-specific fine-tuning yields more accurate and actionable guidance that translates into better editing results.

\begin{table}[t]
\centering
\small
\setlength{\tabcolsep}{3pt}
\begin{tabular}{ccc cccc}
\toprule
Und. & Stage I & Stage II & QAlign$\uparrow$ & DeQA$\uparrow$ & MET3R$\downarrow$ & PF-ass$\uparrow$ \\
\midrule
 & & & 3.272 & 3.869 & 0.257 & 2.987 \\
$\checkmark$ & & & 3.316 & 3.925 & 0.275 & 3.050 \\
$\checkmark$ & $\checkmark$ & & 3.282 & 3.934 & 0.222 & 3.043 \\
$\checkmark$ & $\checkmark$ & $\checkmark$ & \textbf{3.381} & \textbf{4.012} & \textbf{0.184} & \textbf{3.133} \\
\bottomrule
\end{tabular}
\caption{Ablation study of components of GeoComposer.
``Und.'' is the composition understanding model, ``Stage I'' denotes geometry-aware representation learning, and ``Stage II'' denotes hybrid reward-guided reinforcement learning.}
\label{tab:ablation_component}
\end{table}
\begin{table}[t]
\centering
\small
\setlength{\tabcolsep}{4pt}
\begin{tabular}{l cccc}
\toprule
Model & QAlign$\uparrow$ & DeQA$\uparrow$ & MET3R$\downarrow$ & PF-ass$\uparrow$ \\
\midrule
Qwen3-VL & 3.370 & 3.992 & 0.201 & 3.096 \\
PhotoFramer & 3.375 & 3.938 & 0.191 & 3.051 \\
Ours & \textbf{3.381} & \textbf{4.012} & \textbf{0.184} & \textbf{3.133} \\
\bottomrule
\end{tabular}
\caption{Ablation study of composition understanding model.}
\label{tab:ablation_understand}
\end{table}

\noindent\textbf{Geometry-Aware Representation Learning.}
Table~\ref{tab:ablation_stage1} ablates the supervision design in Stage I, evaluated without Stage II to isolate its effect.
Replacing the conventional point-wise L2 supervision with our cosine-based local correspondence supervision improves geometric consistency, and further adding global structural supervision yields the best MET3R (0.222).
These results validate the effectiveness of both local correspondence and global structural supervision.

\noindent\textbf{Hybrid Reward.}
Table~\ref{tab:ablation_reward} ablates each component of the hybrid reward by removing one reward at a time.
Removing the geometry reward yields slightly higher aesthetic scores (QAlign) but causes a severe collapse of geometric consistency, which is a direct manifestation of reward hacking where the policy inflates the editing score through spatially inconsistent modifications.
Conversely, removing the instruction reward attains a marginally lower MET3R, but this stems from under-editing rather than genuine geometric fidelity since the composition and image quality degrade.
Removing the quality reward degrades both DeQA and MET3R.
The full hybrid reward achieves the best overall balance.

\begin{table}[t]
\centering
\small
\setlength{\tabcolsep}{5pt}
\begin{tabular}{l cccc}
\toprule
Supervision & QAlign$\uparrow$ & DeQA$\uparrow$ & MET3R$\downarrow$ & PF-ass$\uparrow$ \\
\midrule
L2 & \underline{3.260} & 3.890 & 0.249 & 3.011 \\
Local & 3.250 & \underline{3.900} & \underline{0.240} & \underline{3.020} \\
Loc.+Glob. & \textbf{3.282} & \textbf{3.934} & \textbf{0.222} & \textbf{3.043} \\
\bottomrule
\end{tabular}
\caption{Ablation study of the supervision design in geometry-aware representation learning, evaluated after Stage I. ``Loc.+Glob.'' denotes local correspondence combined with global structural supervision.}
\label{tab:ablation_stage1}
\end{table}
\begin{table}[t]
\centering
\small
\setlength{\tabcolsep}{4pt}
\begin{tabular}{l cccc}
\toprule
Reward & QAlign$\uparrow$ & DeQA$\uparrow$ & MET3R$\downarrow$ & PF-ass$\uparrow$ \\
\midrule
Full (Ours) & \underline{3.381} & \textbf{4.012} & \underline{0.184} & \textbf{3.133} \\
$-$ Instruction & 3.249 & 3.626 & \textbf{0.182} & 3.012 \\
$-$ Quality & 3.375 & 3.931 & 0.197 & \underline{3.127} \\
$-$ Geometry & \textbf{3.405} & \underline{3.957} & 0.303 & 3.014 \\
\bottomrule
\end{tabular}
\caption{Ablation study of the hybrid reward, where each row removes one reward component from the full hybrid reward.}
\label{tab:ablation_reward}
\end{table}

\section{Conclusion}

In this work, we present a geometry-grounded framework named GeoComposer for multimodal photographic composition.
GeoComposer produces textual guidance from a composition understanding model and generates visual exemplars from a composition editing model with geometry-aware representation learning and hybrid reward-guided reinforcement learning.
Extensive experiments demonstrate that our approach outperforms state-of-the-art methods in generating visually appealing and geometrically consistent compositions.

\section{Limitation and Future Work}

Despite achieving compelling performance, our method still has some limitations.
First, our approach operates on a single input image to generate textual guidance and a visual exemplar.
However, a single view may provide incomplete scene context and fail to capture the full spatial structure of the environment.
Extending the framework to support multiple input views is a promising direction for future work.
In addition, photographic composition is a relatively subjective process, and multiple aesthetically pleasing compositions may exist for the same scene.
By generating only a single visual exemplar, our framework may not fully capture the diversity of individual preferences.
Future work could explore incorporating user-specific preferences to provide more diverse and personalized composition guidance.

{
    \small
    \bibliographystyle{ieeenat_fullname}
    \bibliography{main}

@inproceedings{tu2020image,
  title={Image cropping with composition and saliency aware aesthetic score map},
  author={Tu, Yi and Niu, Li and Zhao, Weijie and Cheng, Dawei and Zhang, Liqing},
  booktitle={Proceedings of the AAAI conference on artificial intelligence},
  volume={34},
  number={07},
  pages={12104--12111},
  year={2020}
}

@article{wu2025qwen,
  title={Qwen-image technical report},
  author={Wu, Chenfei and Li, Jiahao and Zhou, Jingren and Lin, Junyang and Gao, Kaiyuan and Yan, Kun and Yin, Sheng-ming and Bai, Shuai and Xu, Xiao and Chen, Yilei and others},
  journal={arXiv preprint arXiv:2508.02324},
  year={2025}
}

@inproceedings{you2026photoframer,
  title={Photoframer: Multi-modal image composition instruction},
  author={You, Zhiyuan and Wang, Ke and Zhang, He and Cai, Xin and Gu, Jinjin and Xue, Tianfan and Dong, Chao and Zhang, Zhoutong},
  booktitle={Proceedings of the IEEE/CVF Conference on Computer Vision and Pattern Recognition},
  pages={10197--10207},
  year={2026}
}

@article{yu2024representation,
  title={Representation alignment for generation: Training diffusion transformers is easier than you think},
  author={Yu, Sihyun and Kwak, Sangkyung and Jang, Huiwon and Jeong, Jongheon and Huang, Jonathan and Shin, Jinwoo and Xie, Saining},
  journal={arXiv preprint arXiv:2410.06940},
  year={2024}
}

@inproceedings{wang2025vggt,
  title={Vggt: Visual geometry grounded transformer},
  author={Wang, Jianyuan and Chen, Minghao and Karaev, Nikita and Vedaldi, Andrea and Rupprecht, Christian and Novotny, David},
  booktitle={Proceedings of the Computer Vision and Pattern Recognition Conference},
  pages={5294--5306},
  year={2025}
}

@article{yao2026photography,
  title={Photography Perspective Composition: Towards Aesthetic Perspective Recommendation},
  author={Yao, Lujian and Zheng, Siming and Yuan, Xinbin and Cai, Zhuoxuan and Wu, Pu and Chen, Jinwei and Li, Bo and Jiang, Peng-Tao},
  journal={Advances in Neural Information Processing Systems},
  volume={38},
  pages={90306--90327},
  year={2026}
}

@inproceedings{fang2014automatic,
  title={Automatic image cropping using visual composition, boundary simplicity and content preservation models},
  author={Fang, Chen and Lin, Zhe and Mech, Radomir and Shen, Xiaohui},
  booktitle={Proceedings of the 22nd ACM international conference on Multimedia},
  pages={1105--1108},
  year={2014}
}

@article{zeng2020grid,
  title={Grid anchor based image cropping: A new benchmark and an efficient model},
  author={Zeng, Hui and Li, Lida and Cao, Zisheng and Zhang, Lei},
  journal={IEEE Transactions on Pattern Analysis and Machine Intelligence},
  volume={44},
  number={3},
  pages={1304--1319},
  year={2020},
  publisher={IEEE}
}

@inproceedings{hong2021composing,
  title={Composing photos like a photographer},
  author={Hong, Chaoyi and Du, Shuaiyuan and Xian, Ke and Lu, Hao and Cao, Zhiguo and Zhong, Weicai},
  booktitle={Proceedings of the IEEE/CVF Conference on Computer Vision and Pattern Recognition},
  pages={7057--7066},
  year={2021}
}

@article{du2026aesformer,
  title={AesFormer: Transform Everyday Photos into Beautiful Memories},
  author={Du, Tianxiang and He, Hulingxiao and Peng, Yuxin},
  journal={arXiv preprint arXiv:2605.22126},
  year={2026}
}

@article{liu2025step1x,
  title={Step1x-edit: A practical framework for general image editing},
  author={Liu, Shiyu and Han, Yucheng and Xing, Peng and Yin, Fukun and Wang, Rui and Cheng, Wei and Liao, Jiaqi and Wang, Yingming and Fu, Honghao and Han, Chunrui and others},
  journal={arXiv preprint arXiv:2504.17761},
  year={2025}
}

@article{deng2025emerging,
  title={Emerging properties in unified multimodal pretraining},
  author={Deng, Chaorui and Zhu, Deyao and Li, Kunchang and Gou, Chenhui and Li, Feng and Wang, Zeyu and Zhong, Shu and Yu, Weihao and Nie, Xiaonan and Song, Ziang and others},
  journal={arXiv preprint arXiv:2505.14683},
  year={2025}
}

@inproceedings{li2025towards,
  title={Towards Smart Point-and-Shoot Photography},
  author={Li, Jiawan and Zhou, Fei and Zhong, Zhipeng and Lin, Jiongzhi and Qiu, Guoping},
  booktitle={Proceedings of the Computer Vision and Pattern Recognition Conference},
  pages={28242--28251},
  year={2025}
}

@article{singh2025matters,
  title={What matters for Representation Alignment: Global Information or Spatial Structure?},
  author={Singh, Jaskirat and Leng, Xingjian and Wu, Zongze and Zheng, Liang and Zhang, Richard and Shechtman, Eli and Xie, Saining},
  journal={arXiv preprint arXiv:2512.10794},
  year={2025}
}

@article{xu2026beyond,
  title={Beyond Point-Wise Matching: Structural Representation Alignment for Accelerating Diffusion Transformers},
  author={Xu, Shaodong and Wang, Zhendong and Gong, Litong and Li, Zexian and Zhou, Wengang and Ge, Tiezheng and Li, Houqiang},
  journal={arXiv preprint arXiv:2605.16949},
  year={2026}
}

@inproceedings{brooks2023instructpix2pix,
  title={Instructpix2pix: Learning to follow image editing instructions},
  author={Brooks, Tim and Holynski, Aleksander and Efros, Alexei A},
  booktitle={Proceedings of the IEEE/CVF conference on computer vision and pattern recognition},
  pages={18392--18402},
  year={2023}
}

@article{zheng2025diffusionnft,
  title={Diffusionnft: Online diffusion reinforcement with forward process},
  author={Zheng, Kaiwen and Chen, Huayu and Ye, Haotian and Wang, Haoxiang and Zhang, Qinsheng and Jiang, Kai and Su, Hang and Ermon, Stefano and Zhu, Jun and Liu, Ming-Yu},
  journal={arXiv preprint arXiv:2509.16117},
  year={2025}
}

@article{wang2025unified,
  title={Unified reward model for multimodal understanding and generation},
  author={Wang, Yibin and Zang, Yuhang and Li, Hao and Jin, Cheng and Wang, Jiaqi},
  journal={arXiv preprint arXiv:2503.05236},
  year={2025}
}

@article{labs2025flux,
  title={FLUX. 1 Kontext: Flow Matching for In-Context Image Generation and Editing in Latent Space},
  author={Labs, Black Forest and Batifol, Stephen and Blattmann, Andreas and Boesel, Frederic and Consul, Saksham and Diagne, Cyril and Dockhorn, Tim and English, Jack and English, Zion and Esser, Patrick and others},
  journal={arXiv preprint arXiv:2506.15742},
  year={2025}
}

@inproceedings{xiao2025omnigen,
  title={Omnigen: Unified image generation},
  author={Xiao, Shitao and Wang, Yueze and Zhou, Junjie and Yuan, Huaying and Xing, Xingrun and Yan, Ruiran and Li, Chaofan and Wang, Shuting and Huang, Tiejun and Liu, Zheng},
  booktitle={Proceedings of the IEEE/CVF Conference on Computer Vision and Pattern Recognition},
  pages={13294--13304},
  year={2025}
}

@inproceedings{sheynin2024emu,
  title={Emu edit: Precise image editing via recognition and generation tasks},
  author={Sheynin, Shelly and Polyak, Adam and Singer, Uriel and Kirstain, Yuval and Zohar, Amit and Ashual, Oron and Parikh, Devi and Taigman, Yaniv},
  booktitle={Proceedings of the IEEE/CVF Conference on Computer Vision and Pattern Recognition},
  pages={8871--8879},
  year={2024}
}

@article{farhat2022captain,
  title={CAPTAIN: Comprehensive composition assistance for photo taking},
  author={Farhat, Farshid and Kamani, Mohammad Mahdi and Wang, James Z},
  journal={ACM Transactions on Multimedia Computing, Communications, and Applications (TOMM)},
  volume={18},
  number={1},
  pages={1--24},
  year={2022},
  publisher={ACM New York, NY}
}

@article{zhang2018pose,
  title={Pose-based composition improvement for portrait photographs},
  author={Zhang, Xiaoyan and Li, Zhuopeng and Constable, Martin and Chan, Kap Luk and Tang, Zhenhua and Tang, Gaoyang},
  journal={IEEE Transactions on Circuits and Systems for Video Technology},
  volume={29},
  number={3},
  pages={653--668},
  year={2018},
  publisher={IEEE}
}

@article{wang2026pi,
  title={$\pi^3$: Permutation-Equivariant Visual Geometry Learning},
  author={Wang, Yifan and Zhou, Jianjun and Zhu, Haoyi and Chang, Wenzheng and Zhou, Yang and Li, Zizun and Chen, Junyi and Pang, Jiangmiao and Shen, Chunhua and He, Tong},
  journal={arXiv preprint arXiv:2507.13347},
  year={2025}
}

@misc{diffsynthstudio,
  title={DiffSynth-Studio},
  author={{ModelScope}},
  year={2024},
  howpublished={\url{https://github.com/modelscope/DiffSynth-Studio}}
}

@misc{flowfactory,
  title={Flow-Factory},
  author={{X-GenGroup}},
  year={2025},
  howpublished={\url{https://github.com/X-GenGroup/Flow-Factory}}
}

@inproceedings{wu2024qalign,
  title={Q-Align: Teaching LMMs for Visual Scoring via Discrete Text-Defined Levels},
  author={Wu, Haoning and Zhang, Zicheng and Zhang, Weixia and Chen, Chaofeng and Liao, Liang and Li, Chunyi and Gao, Yixuan and Wang, Annan and Zhang, Erli and Sun, Wenxiu and others},
  booktitle={International Conference on Machine Learning},
  year={2024}
}

@inproceedings{you2025deqa,
  title={Teaching Large Language Models to Regress Accurate Image Quality Scores Using Score Distribution},
  author={You, Zhiyuan and Cai, Xin and Gu, Jinjin and Xue, Tianfan and Dong, Chao},
  booktitle={Proceedings of the IEEE/CVF Conference on Computer Vision and Pattern Recognition},
  year={2025}
}

@inproceedings{asim2025met3r,
  title={MEt3R: Measuring Multi-View Consistency in Generated Images},
  author={Asim, Mohammad and Wewer, Christopher and Wimmer, Thomas and Schiele, Bernt and Lenssen, Jan Eric},
  booktitle={Proceedings of the IEEE/CVF Conference on Computer Vision and Pattern Recognition},
  year={2025}
}

@inproceedings{ling2024dl3dv,
  title={DL3DV-10K: A Large-Scale Scene Dataset for Deep Learning-Based 3D Vision},
  author={Ling, Lu and Sheng, Yichen and Tu, Zhi and Zhao, Wentian and Xin, Cheng and Wan, Kun and Yu, Lantao and Guo, Qianyu and Yu, Zixun and Lu, Yawen and others},
  booktitle={Proceedings of the IEEE/CVF Conference on Computer Vision and Pattern Recognition},
  pages={22160--22169},
  year={2024}
}

@article{qwen3vl2025,
  title={Qwen3-VL Technical Report},
  author={{Qwen Team}},
  journal={arXiv preprint arXiv:2511.21631},
  year={2025}
}

@misc{labs2025flux2,
  title={FLUX.2: Frontier Visual Intelligence},
  author={{Black Forest Labs}},
  howpublished={\url{https://bfl.ai/blog/flux-2}},
  year={2025}
}

@misc{google2025nanobananapro,
  title={Nano Banana Pro (Gemini 3 Pro Image)},
  author={{Google DeepMind}},
  howpublished={\url{https://deepmind.google/models/gemini-image/pro/}},
  year={2025}
}

@article{radenovic2018fine,
  title={Fine-tuning {CNN} Image Retrieval with No Human Annotation},
  author={Radenovi{\'c}, Filip and Tolias, Giorgos and Chum, Ond{\v{r}}ej},
  journal={IEEE Transactions on Pattern Analysis and Machine Intelligence},
  volume={41},
  number={7},
  pages={1655--1668},
  year={2019},
  publisher={IEEE}
}

@inproceedings{lipman2022flow,
  title={Flow Matching for Generative Modeling},
  author={Lipman, Yaron and Chen, Ricky T. Q. and Ben-Hamu, Heli and Nickel, Maximilian and Le, Matthew},
  booktitle={The Eleventh International Conference on Learning Representations},
  year={2023}
}

@inproceedings{wei2018good,
  title={Good View Hunting: Learning Photo Composition from Dense View Pairs},
  author={Wei, Zijun and Zhang, Jianming and Shen, Xiaohui and Lin, Zhe and M{\v{e}}ch, Radom{\'\i}r and Hoai, Minh and Samaras, Dimitris},
  booktitle={Proceedings of the IEEE Conference on Computer Vision and Pattern Recognition},
  pages={2496--2505},
  year={2018}
}

@inproceedings{yan2013learning,
  title={Learning the Change for Automatic Image Cropping},
  author={Yan, Jianzhou and Lin, Stephen and Kang, Sing Bing and Tang, Xiaoou},
  booktitle={Proceedings of the IEEE Conference on Computer Vision and Pattern Recognition},
  pages={971--978},
  year={2013}
}

@inproceedings{radford2021learning,
  title={Learning Transferable Visual Models from Natural Language Supervision},
  author={Radford, Alec and Kim, Jong Wook and Hallacy, Chris and Ramesh, Aditya and Goh, Gabriel and Agarwal, Sandhini and Sastry, Girish and Askell, Amanda and Mishkin, Pamela and Clark, Jack and others},
  booktitle={International Conference on Machine Learning},
  pages={8748--8763},
  year={2021}
}

@article{oquab2023dinov2,
  title={DINOv2: Learning Robust Visual Features without Supervision},
  author={Oquab, Maxime and Darcet, Timoth{\'e}e and Moutakanni, Th{\'e}o and Vo, Huy and Szafraniec, Marc and Khalidov, Vasil and Fernandez, Pierre and Haziza, Daniel and Massa, Francisco and El-Nouby, Alaaeldin and others},
  journal={Transactions on Machine Learning Research},
  year={2024}
}

@article{qin2020u2net,
  title={U$^2$-Net: Going Deeper with Nested U-Structure for Salient Object Detection},
  author={Qin, Xuebin and Zhang, Zichen and Huang, Chenyang and Dehghan, Masood and Zaiane, Osmar R and Jagersand, Martin},
  journal={Pattern Recognition},
  volume={106},
  pages={107404},
  year={2020},
  publisher={Elsevier}
}

@misc{unsplash,
  title={Unsplash Lite Dataset},
  author={{Unsplash}},
  year={2023},
  howpublished={\url{https://github.com/unsplash/datasets}}
}

@article{han2025emergent,
  title={Emergent Outlier View Rejection in Visual Geometry Grounded Transformers},
  author={Han, Jisang and Hong, Sunghwan and Jung, Jaewoo and Jang, Wooseok and
          An, Honggyu and Wang, Qianqian and Kim, Seungryong and Feng, Chen},
  journal={arXiv preprint arXiv:2512.04012},
  year={2025}
}
}

\twocolumn[{%
\centering
{\Large\bfseries Supplementary Material for\\[4pt]
GeoComposer: Geometry-Grounded Photographic Composition Instruction\par}
\vspace{2em}
}]

\appendix

In this supplementary material, we provide additional details on
(1) Dataset Construction,
(2) Evaluation Metrics,
(3) Baseline Methods,
(4) Implementation Details,
(5) Algorithmic Pseudocode,
(6) Results on a Held-out Benchmark,
(7) Additional Qualitative Results,
(8) Analysis of the Composition Understanding Model, and
(9) Additional Ablation Studies.

\section{Dataset Construction}
\label{sec:dataset_construction}

We follow the data construction protocol of PhotoFramer~\cite{you2026photoframer} and build the dataset on top of existing public datasets, i.e., the human-annotated crop-scoring datasets and expert photographs listed below, from which the composition pairs and their text guidance are derived.
The constructed dataset consists of 34,129 $\langle$\emph{poor}, \emph{good}, \emph{text-guidance}$\rangle$ triplets organized into three composition operations that mirror how casual photographs are re-composed, namely \emph{shift} (in-plane re-framing), \emph{zoom-in} (tightening the framing toward the subject), and \emph{view change} (camera viewpoint variation).
Each triplet pairs a poorly-composed image with a well-composed counterpart of the same scene, together with a natural-language guidance describing the poor$\rightarrow$good compositional adjustment.
Table~\ref{tab:dataset_stats} summarizes the per-task statistics.
The \emph{shift} and \emph{zoom-in} pairs are obtained by re-cropping real photographs, whereas the \emph{view change} pairs are synthesized from expert photographs; all accepted pairs are finally annotated with text guidance.

\noindent\textbf{Shift and Zoom-in Pairs.}
The two subsets are built from crop-scoring datasets, i.e., CPC~\cite{wei2018good}, GAIC~\cite{zeng2020grid}, FLMS~\cite{fang2014automatic}, and CUHK-ICD~\cite{yan2013learning}, whose expert crops serve as well-composed references, with the per-subset sources listed in Table~\ref{tab:dataset_stats}.
For \emph{shift}, we sample two crops of the same aspect ratio from a source image, where the \emph{good} crop is a high-scoring expert crop and the \emph{poor} crop is a positionally offset crop, to which we further apply a small random rotation to emulate the tilted horizon typical of amateur framing.
For \emph{zoom-in}, the \emph{good} image is a tightly composed high-scoring crop and the \emph{poor} image is a larger crop that contains it, emulating a wider or more distant framing; we enforce that the good crop covers at most 60\% of the poor crop in area to ensure a genuine change of framing rather than a near-identical re-crop.
A candidate pair passes a cascade of filters that jointly enforce three properties, and is accepted only if all of them are satisfied.
(i) \emph{Same scene}: the poor and good crops must depict the same content, verified by a CLIP~\cite{radford2021learning} similarity $\geq 0.8$ and a subject-consistency check that matches U$^2$-Net~\cite{qin2020u2net} salient regions via DINOv2~\cite{oquab2023dinov2} cosine similarity $\geq 0.6$.
(ii) \emph{Composition gap}: the good crop must be genuinely better composed, enforced by an aesthetic gate on the good crop (Q-Align~\cite{wu2024qalign} $\geq 2.5$) and a positive composition-score margin between the good and poor crops given by the source annotations.
(iii) \emph{Valid operation}: we pair only crops of the same aspect ratio within $[0.45, 2.2]$, and for \emph{shift} we additionally reject pairs in which one crop nearly contains the other, as such a containment relation corresponds to a zoom rather than an in-plane shift.

\begin{table}[t]
\centering
\small
\setlength{\tabcolsep}{5pt}
\begin{tabular}{l c l}
\toprule
Subset & Pairs & Source \\
\midrule
Shift & 13,152 & CPC, GAIC \\
Zoom-in & 10,454 & CPC, FLMS, GAIC, CUHK-ICD \\
View change & 10,523 & Unsplash \\
\midrule
Total & 34,129 & -- \\
\bottomrule
\end{tabular}
\caption{Per-task statistics of the constructed dataset. The view-change poor images are synthesized from Unsplash expert photographs.}
\label{tab:dataset_stats}
\end{table}

\noindent\textbf{View-Change Pairs.}
Paired real viewpoint-change data are scarce, and their aesthetic and compositional quality is inherently limited, e.g., multi-view frames captured for 3D reconstruction, which caps the quality of the frames that can serve as well-composed \emph{good} references.
Since the \emph{good} image defines the compositional quality the model is trained to reproduce, it must itself satisfy a high aesthetic and compositional standard.
We therefore synthesize this subset from a curated dataset of real high-quality photographs, Unsplash~\cite{unsplash}. 
Starting from an expert photograph as the \emph{good} image, we apply camera-motion editing instructions (e.g., turn, move, rotate) to different directions with discrete magnitudes (slightly, moderately, and strongly) with Qwen-Image-Edit~\cite{wu2025qwen} to generate a viewpoint-shifted \emph{poor} image with explicit prompts of worsening composition.
We then retain only genuine and correctly degraded pairs through a three-axis filter:
(1) \emph{Same Scene}: besides a CLIP similarity $\geq 0.8$ and a DINOv2 cosine similarity $\geq 0.6$ between the salient regions of the poor and good images, we also leverage Qwen3.5-397B-A17B and human annotators to verify that the poor image is indeed a viewpoint-changed version of the good one;
(2) \emph{Valid Discrimination}: a viewpoint-shift test based on mutual nearest-neighbor matching of DINOv2 patch tokens, keeping only pairs whose spatial displacement exceeds 5\%, and a non-mirror test that removes pairs where the poor image is a left-right flip of the good one;
and (3) \emph{Composition gap}: a non-inversion test that removes pairs whose synthesized ``poor'' image is in fact better composed than the good one, according to the Q-Align and PhotoFramer composition scorers~\cite{you2026photoframer}.

\noindent\textbf{Text-Guidance Annotation.}
For every $\langle$poor, good$\rangle$ pair, we generate a natural-language editing instruction with a vision-language model (Qwen3.5-397B-A17B), which is shown the poor image as ``Input photo'' and the good image as ``Desired result'' and, cast as an expert composition editor, returns a self-contained instruction describing how to reframe the former into the latter.
The prompt constrains the model to composition-only edits (subject placement, balance, horizon, viewpoint) and forbids introducing new content or numeric coordinates, keeping each instruction grounded in the pair and free of hallucination. 

\noindent\textbf{Training and Validation Split.}
We split the dataset into 33,229 training and 900 validation triplets (300 per task).
The split is strictly disjoint at the source-image level so that no source image appears in both splits, preventing information leakage between training and evaluation.

\noindent\textbf{Held-out Benchmark.}
To assess generalization beyond the constructed dataset, we additionally evaluate on a held-out benchmark that is disjoint from our training data and carries no text guidance, i.e., pure image-conditioned re-composition.
For \emph{view change}, we sample 300 real multi-view pairs from DL3DV~\cite{ling2024dl3dv}, where the poor image is a scene's native first frame and the good image is a better-composed frame (scored by the PhotoFramer composition scorer) within co-visibility $[0.1, 0.3]$ with the poor image, providing a fully real out-of-distribution test of viewpoint change.

\section{Evaluation Metrics}
\label{sec:metric_details}
We evaluate composition editing along four orthogonal axes, i.e., composition, image quality, geometric consistency, and text-guidance accuracy, together with a human study.
Unless otherwise noted, all LLM-judge metrics use the same frontier vision-language model (GPT-5.4) as the judge, with input images normalized to a long side of 640 pixels to remove resolution bias.

\noindent\emph{Composition.}
\underline{Comp.~WinRate} is the pairwise win rate of the edited image against the source, where the judge decides which image is better composed.
We adopt double-order judging that averages both presentation orders to remove position bias, so that a value above 50\% indicates improved composition while a near-identity edit lands around 50\%.
\underline{PF-ass} scores single-image composition with the PhotoFramer scorer~\cite{you2026photoframer}, with images downsampled to a 0.3M-pixel cap to remove its systematic resolution bias.

\noindent\emph{Image Quality.}
\underline{QAlign}~\cite{wu2024qalign} scores no-reference aesthetic quality with the aesthetics head of Q-Align, and \underline{DeQA}~\cite{you2025deqa} scores general quality in terms of distortion, artifacts, and naturalness.

\noindent\emph{Geometric Consistency.}
\underline{MET3R}~\cite{asim2025met3r} measures multi-view 3D consistency between the edited output and the source, where lower values indicate greater agreement.
Since MET3R inherently rewards under-editing, we complement it with \underline{Geo-Suc}, a pointwise LLM-judge metric that takes the reference camera motion defined by the source-to-target pair and decides whether the source-to-output change is explainable by the same motion.
It is calibrated against three control arms, i.e., the ground truth as its own candidate (ceiling), the source itself (floor), and a shuffled scene (sanity).

\noindent\emph{Text-Guidance Accuracy.}
Following the protocol of PhotoFramer~\cite{you2026photoframer}, \underline{Text Cons.~(src$\rightarrow$gen)} scores whether the reasoning text accurately describes the method's own edit, and \underline{Text Cons.~(src$\rightarrow$GT)} whether it matches the human reframing, both on a hedging-resistant 1--5 rubric that places unfalsifiable or generic guidance at the bottom.
Pure image generators that emit no reasoning text are excluded from these two metrics.

\noindent\emph{Human Study.}
On a randomly sampled 10\% subset of the validation set, annotators are shown the source and an edited candidate and judge whether the candidate, while satisfying geometric consistency, improves composition and image quality over the source.
We report \underline{Overall}, the fraction of samples for which a method's edit is judged both geometrically valid and better than the source in composition and quality; higher is better.

\section{Baseline Details}
\label{sec:baseline_details}
Under a unified protocol, every model, including ours, receives the general task prompts of the dataset as the textual input together with the poorly-composed image, and produces a composition-improved image with, depending on its capability, an accompanying text guidance.
BAGEL~\cite{deng2025emerging} is an open-source unified multimodal model that couples understanding and generation within a single architecture.
Qwen-Image-Edit~\cite{wu2025qwen} shares the same editing backbone as ours and is paired with Qwen3-VL~\cite{qwen3vl2025} as its understanding model.
Step1X-Edit~\cite{liu2025step1x} is a general instruction-based editing framework, which we run in its thinking mode that produces an explicit reasoning trace before editing.
FLUX.2~\cite{labs2025flux2} is an open-weight flow-matching model for high-quality image generation and editing.
PhotoFramer~\cite{you2026photoframer} is a composition-specific unified understanding-generation model that jointly produces composition guidance and visual exemplars, and we adopt its released preview version.
Nano Banana Pro (Gemini 3 Pro Image)~\cite{google2025nanobananapro} is a proprietary frontier image generation and editing model.
Among these, all methods except FLUX.2 and Nano Banana Pro emit textual output that serves as the text guidance, whereas the two pure image generators produce only the edited image.

\begin{algorithm}[t]
\caption{Composition Understanding}
\label{alg:understanding}
\begin{algorithmic}[1]

\Require Poorly-composed image $I^p$ and task prompt $T_{task}$
\Ensure Textual composition guidance $T$

\State Load the fine-tuned composition understanding model $f_{und}$
\State Generate the textual guidance $T \leftarrow f_{und}(I^p, T_{task})$
\State \Return $T$

\end{algorithmic}
\end{algorithm}

\begin{algorithm}[t]
\caption{Training Stage I (Geometry-Aware Representation Learning)}
\label{alg:geometry_training}
\begin{algorithmic}[1]

\Require Dataset $\mathcal{D}=\{(I^p,I^t,T)\}$ and number of epochs $E$
\Ensure Trained DiT $f_{\mathrm{DiT}}$

\State Initialize the DiT $f_{\mathrm{DiT}}$ and the projection head $P$
\State Load the frozen geometry foundation model $f_{geo}$

\For{$e=1$ to $E$}
\ForAll{$(I^p,I^t,T)\in\mathcal{D}$}

    \State Extract $\{F^{\mathrm{patch}},F^{\mathrm{cam}},F^{\mathrm{reg}}\}\leftarrow f_{geo}(I^t)$

    \State Build the multimodal condition $c$ from $T$ and $I^p$

    \State Encode $z_0 \leftarrow \mathrm{VAE}(I^t)$ and concatenate the
    special tokens $\tilde{z}_0 \leftarrow [\,z_0;F^{\mathrm{cam}};F^{\mathrm{reg}}\,]$

    \State Sample $t$ and $\epsilon\sim\mathcal{N}(0,I)$, and add noise to
    $\tilde{z}_0$ to obtain $\tilde{z}_t$

    \State Predict $(\hat{v},F^{\mathrm{DiT}})\leftarrow f_{\mathrm{DiT}}(\tilde{z}_t,t,c)$
    and project $\hat{F}^{\mathrm{DiT}} \leftarrow P(F^{\mathrm{DiT}})$

    \State $\mathcal{L}_{\mathrm{geo}}
    \leftarrow
    \mathcal{L}_{\mathrm{FM}}
    +
    \lambda_{\mathrm{glo}}\mathcal{L}_{\mathrm{glo}}
    +
    \lambda_{\mathrm{loc}}\mathcal{L}_{\mathrm{loc}}$

    \State Update $f_{\mathrm{DiT}}$ and $P$ using $\mathcal{L}_{\mathrm{geo}}$

\EndFor
\EndFor

\State \Return Trained DiT $f_{\mathrm{DiT}}$

\end{algorithmic}
\end{algorithm}

\begin{algorithm}[t]
\caption{Training Stage II (Hybrid Reward-Guided RL)}
\label{alg:reinforcement_learning}
\begin{algorithmic}[1]

\Require Dataset $\mathcal{D}=\{(I^p,I^t,T)\}$, rollout size $K$, and epochs $E'$
\Ensure Hybrid reward-guided RL-optimized DiT

\State Load the DiT trained from Stage~I and initialize the policy
$\pi_{\theta}$ and the frozen reference policy $\pi_{\mathrm{ref}}$ from it

\For{$e=1$ to $E'$}
\ForAll{$(I^p,I^t,T)\in\mathcal{D}$}

    \State Rollout $K$ candidates
    $\mathcal{Y}=\{I^c_1,\dots,I^c_K\}\sim\pi_{\theta}(I^p,T)$

    \For{$k=1$ to $K$}
        \State $R_{\mathrm{hyb}}^k
        \leftarrow
        \lambda_{\mathrm{ins}}R_{\mathrm{ins}}^k
        +
        \lambda_{\mathrm{qua}}R_{\mathrm{qua}}^k
        +
        \lambda_{\mathrm{geo}}R_{\mathrm{geo}}^k$
    \EndFor

    \State Compute group-relative advantages $\{A^k\}_{k=1}^{K}$
    by normalizing $\{R_{\mathrm{hyb}}^k\}_{k=1}^{K}$ within the group

    \State Update $\pi_{\theta}$ with the negative-aware flow-matching
    objective, weighting the $\pi_{\mathrm{ref}}$-mirrored negative
    targets by $\{A^k\}_{k=1}^{K}$

\EndFor
\EndFor

\State \Return Optimized DiT $\pi_{\theta}$

\end{algorithmic}
\end{algorithm}

\begin{algorithm}[t]
\caption{Inference}
\label{alg:inference}
\begin{algorithmic}[1]

\Require Poorly-composed image $I^p$ and task prompt $T_{task}$
\Ensure Textual guidance $T$ and composition-enhanced exemplar $I^g$

\State Load the understanding model $f_{und}$ and the two-stage trained DiT $f_{\mathrm{DiT}}$
\State Generate the textual guidance $T \leftarrow f_{und}(I^p,T_{task})$
\State Replace the special tokens $F^{\mathrm{cam}}$ and $F^{\mathrm{reg}}$ with Gaussian noise
\State Generate the exemplar $I^{g}\leftarrow f_{\mathrm{DiT}}(I^p,T)$
\State \Return $T,\ I^{g}$

\end{algorithmic}
\end{algorithm}

\section{Implementation Details}
\label{sec:impl_details}
Our composition editing model builds upon Qwen-Image-Edit-2511~\cite{wu2025qwen}, fine-tuned with LoRA of rank 32, while our composition understanding model is obtained by LoRA fine-tuning Qwen3-VL-32B-Instruct~\cite{qwen3vl2025} with a rank of 32 on the constructed dataset.
Following PhotoFramer~\cite{you2026photoframer}, we design ten different general task prompts $T_{task}$ for each task, so that the understanding model is not tied to a single phrasing.
The two stages are implemented with DiffSynth-Studio~\cite{diffsynthstudio} and Flow-Factory~\cite{flowfactory}, and all training is conducted on 8 GPUs with 80 GB memory each.
In Stage~I, we adopt the frozen VGGT-1B~\cite{wang2025vggt} as the geometry foundation model, whose 24-layer aggregator produces 2048-dimensional geometry-aware tokens.
We tap the intermediate image-token features from the 30-th DiT transformer block and align them, through the projection head, to the VGGT patch features at layers $\{12,18,23\}$ for the local correspondence and global structural supervision, while the camera and register tokens are taken from layers $\{18,23\}$ and concatenated with the target latent $z_0$ before noise is added during training.
The supervision weights are $\lambda_{\mathrm{loc}}=0.01$ and $\lambda_{\mathrm{glo}}=0.1$, and the model is optimized with AdamW at a learning rate of $1\times10^{-4}$ for 5 epochs.
In Stage~II, we generate $K=16$ candidates per input through rollout sampling and train for 240 epochs with 48 samples per epoch.
The instruction-following and visual-quality rewards are provided by UnifiedReward~\cite{wang2025unified} and the geometry-consistency reward by Pi3~\cite{wang2026pi} with its patch and attention features probed from layer 17.
The reward weights are $\lambda_{\mathrm{ins}}=\lambda_{\mathrm{qua}}=\lambda_{\mathrm{geo}}=0.5$, the source and target weights $w_s=0.3$ and $w_t=0.7$, and the attention and feature relevance weights $\lambda_{\mathrm{attn}}=0.3$ and $\lambda_{\mathrm{feat}}=0.7$.
The policy is optimized with AdamW at $1\times10^{-4}$, and inference uses 40 diffusion sampling steps.

\section{Algorithms}
\label{sec:algorithms}
We summarize the full GeoComposer pipeline in Algorithms~\ref{alg:understanding}--\ref{alg:inference}, covering composition understanding, geometry-aware representation learning (Stage~I), hybrid reward-guided reinforcement learning (Stage~II), and inference.

\section{Results on a Held-out Benchmark}
\label{sec:public_benchmarks}

\begin{figure*}[t]
\centering
\includegraphics[width=\textwidth]{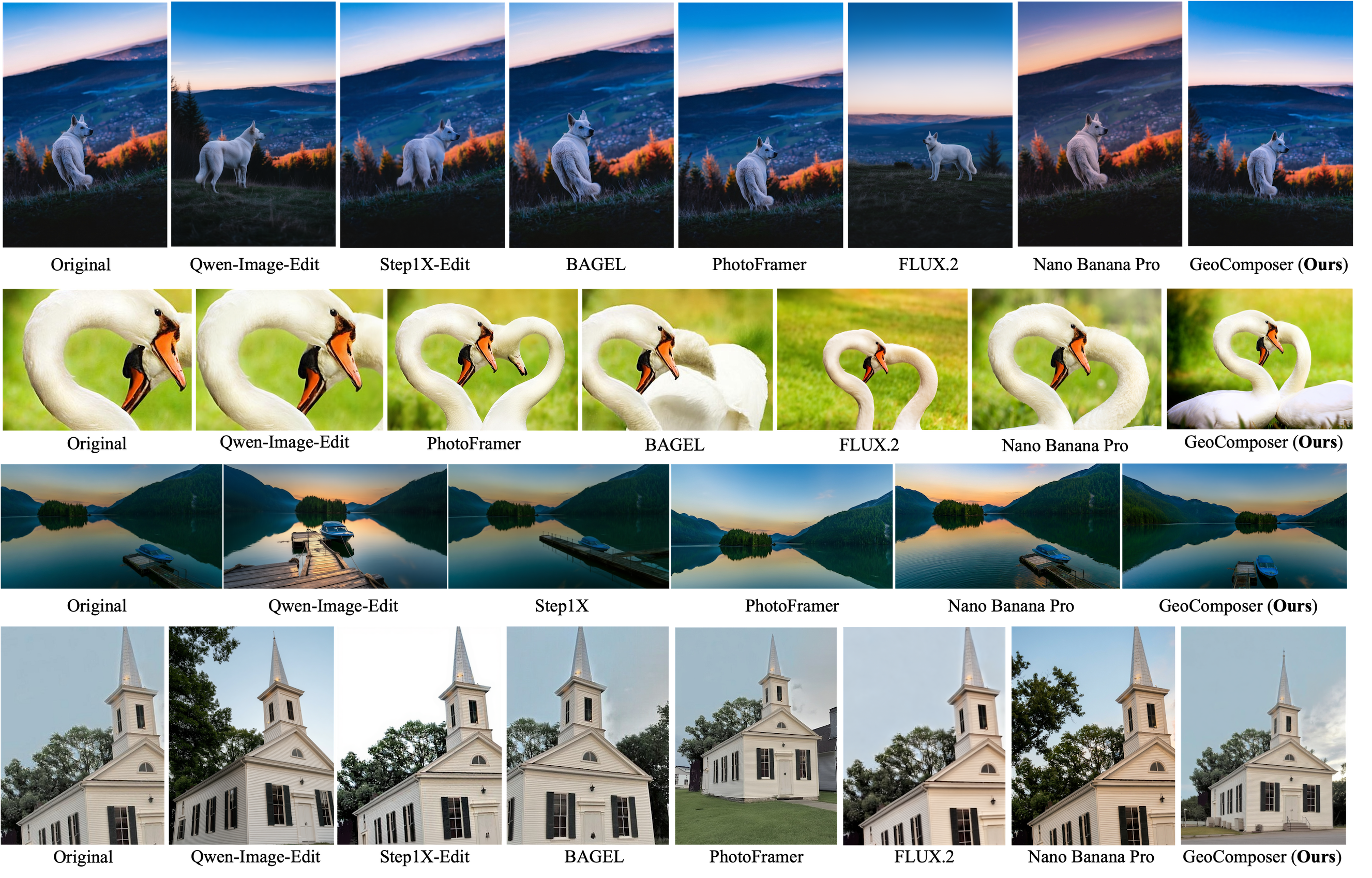}
\caption{Additional qualitative comparison with state-of-the-art methods on diverse scenes. In each row, the leftmost image is the poorly-composed input and the rightmost is the result of GeoComposer.}
\label{fig:qualitative_supp}
\end{figure*}

To assess generalization beyond the constructed dataset, we evaluate on a held-out benchmark that is strictly disjoint from our training data, i.e., DL3DV~\cite{ling2024dl3dv} for viewpoint change, under pure image-conditioned re-composition without text guidance.
Table~\ref{tab:dl3dv} reports the results on the DL3DV-300 subset.
On DL3DV, which requires genuine viewpoint change, GeoComposer attains the best MET3R (0.150) together with the best QAlign, DeQA, and text-guidance faithfulness, whereas several strong editors collapse in geometric consistency, e.g., PhotoFramer reaches a MET3R of 0.453 as it alters the scene without preserving its 3D structure.
GeoComposer is therefore the only method that ranks first on geometric consistency on this benchmark, confirming that the geometric priors learned on the constructed dataset transfer to unseen real data rather than overfitting to our synthesis pipeline.

\begin{table}[t]
\centering
\footnotesize
\setlength{\tabcolsep}{2pt}
\begin{tabular}{l ccccc}
\toprule
Method & QAlign$\uparrow$ & DeQA$\uparrow$ & MET3R$\downarrow$ & PF-ass$\uparrow$ & Cons.$\uparrow$ \\
\midrule
BAGEL & 2.208 & 4.063 & 0.266 & 2.452 & 2.263 \\
Qwen-Image-Edit & \underline{2.323} & 4.016 & 0.242 & 2.456 & 3.337 \\
Step1X-Edit & 2.087 & 3.777 & 0.231 & 2.327 & 2.407 \\
FLUX.2 & 2.247 & 4.027 & 0.259 & 2.563 & -- \\
PhotoFramer & 2.299 & \underline{4.182} & 0.453 & \textbf{2.698} & \underline{3.423} \\
Nano Banana Pro & 2.277 & 4.082 & \underline{0.177} & 2.470 & -- \\
\midrule
Ours & \textbf{2.428} & \textbf{4.253} & \textbf{0.150} & \underline{2.567} & \textbf{3.453} \\
\bottomrule
\end{tabular}
\caption{Comparison on the DL3DV-300 benchmark for viewpoint change.
Cons.~denotes Text Cons.~(src$\rightarrow$gen); ``--'' denotes pure image generators excluded from text-guidance evaluation.
The best and second-best results are marked in \textbf{bold} and \underline{underline}, respectively.}
\label{tab:dl3dv}
\end{table}

\section{Additional Qualitative Results}
\label{sec:additional_qualitative}

Figure~\ref{fig:qualitative_supp} presents additional qualitative comparisons that complement the qualitative results in the main paper, covering a broader range of scene types, i.e., animal subjects, natural landscapes, and architectural structures, across the three composition operations of the dataset.
Consistent with the quantitative results, the baselines frequently improve the framing at the expense of the underlying scene, e.g., hallucinating or removing structures, distorting the subject, or drifting to a viewpoint that is no longer explainable by a camera motion over the source scene.
GeoComposer instead re-frames the subject and rebalances the layout while keeping the scene geometry intact, producing visual exemplars that are both better composed and faithful to the source scene.

\begin{figure*}[t]
\centering
\includegraphics[width=\textwidth]{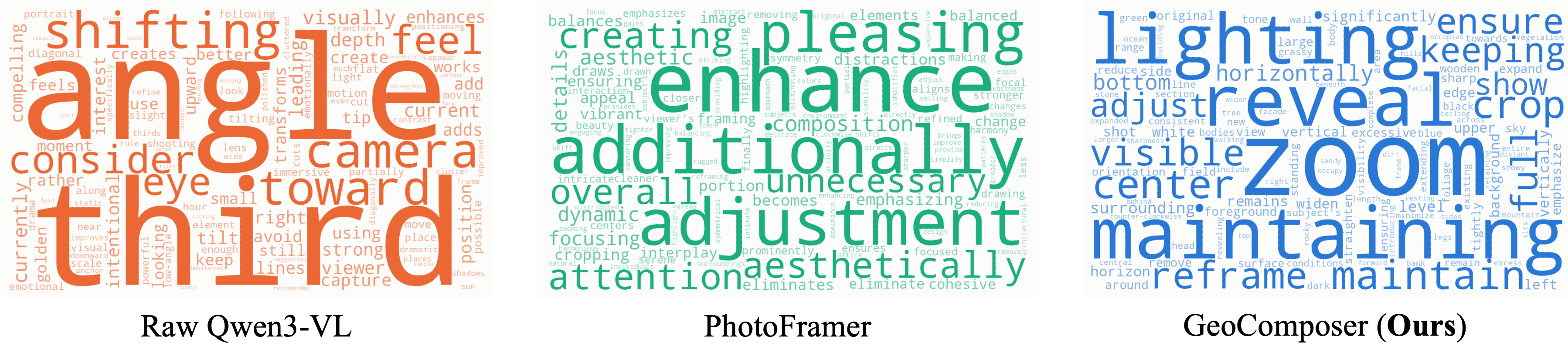}
\caption{Word distribution of the textual guidance produced by each composition understanding model over the validation set.
Word size is proportional to frequency.}
\label{fig:word_und}
\end{figure*}

\section{Analysis of the Composition Understanding Model}
\label{sec:understanding_analysis}

The two components of GeoComposer are evaluated separately in this section, focusing on the textual guidance $T$ rather than the generated exemplar.
Table~\ref{tab:understanding} compares our fine-tuned understanding model against the raw Qwen3-VL~\cite{qwen3vl2025} used by Qwen-Image-Edit and against PhotoFramer~\cite{you2026photoframer}, reporting both text-consistency metrics together with the average length of the produced guidance.
Our model attains the best consistency in both directions (3.45 and 3.28) while producing by far the shortest guidance, i.e., 58.6 words against 100.9 for PhotoFramer and 210.6 for the raw VLM.
The improvement therefore does not come from describing more, but from describing the right thing: length and faithfulness move in opposite directions here, and the longest outputs are the least consistent with the actual reframing.

Figure~\ref{fig:word_und} makes the nature of this difference visible by showing the word distribution of each model over the validation set.
The raw VLM is dominated by composition \emph{theory} and subjective impression, e.g., ``angle'', ``third'', ``diagonal'', and ``golden'' alongside ``feel'' and ``compelling'', so it names the principle a photograph violates and how the result should feel, but rarely states an operation.
PhotoFramer instead concentrates on generic evaluative language, e.g., ``pleasing'', ``aesthetically'', ``enhance'', and ``adjustment'', which reads fluently as composition commentary yet leaves the required change under-specified.
Our guidance is dominated by explicit operations, e.g., ``zoom'', ``crop'', ``reframe'', and ``adjust'', anchored to spatial referents, e.g., ``horizontally'', ``horizon'', ``center'', ``foreground'', and ``background''.
Notably, preservation verbs such as ``maintaining'', ``keeping'', and ``ensure'' are also prominent, which means the guidance states not only what to change but what must stay fixed.
This is precisely the property that makes $T$ usable as a conditioning signal for geometry-preserving editing, and it explains why a much shorter instruction transfers better to the editing model.

\begin{table}[t]
\centering
\small
\setlength{\tabcolsep}{4pt}
\begin{tabular}{l ccc}
\toprule
Und.~Model & src$\rightarrow$gen$\uparrow$ & src$\rightarrow$GT$\uparrow$ & Words \\
\midrule
Raw Qwen3-VL & 3.16 & 2.60 & 210.6 \\
PhotoFramer & \underline{3.21} & \underline{2.69} & 100.9 \\
Ours & \textbf{3.45} & \textbf{3.28} & 58.6 \\
\bottomrule
\end{tabular}
\caption{Comparison of composition understanding models on text consistency and the average number of words per generated guidance.
The best and second-best consistency scores are marked in \textbf{bold} and \underline{underline}, respectively.}
\label{tab:understanding}
\end{table}

\section{Additional Ablation Studies}
\label{sec:additional_ablation}

We provide additional ablations on the design choices of geometry-aware representation learning (Stage~I) and the geometry-consistency reward (Stage~II).
All variants are evaluated on the constructed validation set with QAlign, DeQA, MET3R, and PF-ass.

\begin{table}[t]
\centering
\small
\setlength{\tabcolsep}{5pt}
\begin{tabular}{l cccc}
\toprule
$\lambda_{\mathrm{glo}}$ & QAlign$\uparrow$ & DeQA$\uparrow$ & MET3R$\downarrow$ & PF-ass$\uparrow$ \\
\midrule
0.05 & \textbf{3.310} & 3.908 & 0.242 & 3.022 \\
0.10 & \underline{3.282} & \textbf{3.934} & \textbf{0.222} & \textbf{3.043} \\
0.20 & 3.274 & \underline{3.926} & \underline{0.235} & \underline{3.038} \\
\bottomrule
\end{tabular}
\caption{Ablation of the global structural supervision weight $\lambda_{\mathrm{glo}}$.}
\label{tab:abl_lambda}
\end{table}

\noindent\textbf{Global Supervision Weight $\lambda_{\mathrm{glo}}$.}
Table~\ref{tab:abl_lambda} varies the weight of the global structural supervision.
A small weight ($\lambda_{\mathrm{glo}}=0.05$) attains the best QAlign but leaves geometric consistency weak (MET3R 0.242) and scores worst on DeQA and PF-ass, as the global geometry signal is under-weighted, whereas a large weight ($0.20$) over-constrains the features and improves neither quality nor geometry.
Setting $\lambda_{\mathrm{glo}}=0.1$ attains the best MET3R (0.222) together with the best DeQA and PF-ass, striking the best balance between generation quality and geometric consistency.
As our objective is geometry-grounded composition, we adopt $\lambda_{\mathrm{glo}}=0.1$.

\begin{table}[t]
\centering
\small
\setlength{\tabcolsep}{5pt}
\begin{tabular}{l cccc}
\toprule
Layers & QAlign$\uparrow$ & DeQA$\uparrow$ & MET3R$\downarrow$ & PF-ass$\uparrow$ \\
\midrule
23 & \underline{3.267} & \underline{3.933} & 0.256 & 3.038 \\
18,23 & 3.204 & 3.926 & \underline{0.241} & \textbf{3.068} \\
12,18,23 & \textbf{3.282} & \textbf{3.934} & \textbf{0.222} & \underline{3.043} \\
\bottomrule
\end{tabular}
\caption{Ablation of the VGGT layers whose patch features serve as the target of the local and global supervision in Stage~I.}
\label{tab:abl_sup_layers}
\end{table}

\noindent\textbf{Supervision Layers (Stage~I).}
Table~\ref{tab:abl_sup_layers} varies which VGGT layers provide the patch features that the projected DiT representations are aligned to, i.e., the targets of the local correspondence and global structural supervision.
Supervising against a single deep layer ($\{23\}$) or a single pair ($\{18,23\}$) leaves MET3R at 0.256 and 0.241, as the geometric target is drawn from too narrow a range of depths.
Spanning three depths ($\{12,18,23\}$) attains the best MET3R (0.222) together with the best QAlign and DeQA at a small cost in PF-ass, and we adopt it as our default.

\begin{table}[t]
\centering
\small
\setlength{\tabcolsep}{5pt}
\begin{tabular}{l cccc}
\toprule
$\lambda_{\mathrm{attn}}{:}\lambda_{\mathrm{feat}}$ & QAlign$\uparrow$ & DeQA$\uparrow$ & MET3R$\downarrow$ & PF-ass$\uparrow$ \\
\midrule
0:1 & 3.294 & 3.990 & \underline{0.185} & 3.063 \\
3:7 & \textbf{3.381} & \textbf{4.012} & \textbf{0.184} & \underline{3.133} \\
5:5 & \underline{3.377} & \underline{4.006} & 0.206 & \textbf{3.143} \\
\bottomrule
\end{tabular}
\caption{Ablation of the attention and feature relevance weights $\lambda_{\mathrm{attn}}{:}\lambda_{\mathrm{feat}}$ in the directed geometry relevance score.}
\label{tab:abl_attnfeat}
\end{table}

\noindent\textbf{Attention-Feature Balance (Stage~II).}
Table~\ref{tab:abl_attnfeat} varies the weights $\lambda_{\mathrm{attn}}{:}\lambda_{\mathrm{feat}}$ that combine the attention relevance and the feature similarity into the directed geometry relevance score.
Discarding the attention cue entirely ($0{:}1$) preserves MET3R almost unchanged (0.185) but loses noticeably on composition and quality (PF-ass 3.063, QAlign 3.294), indicating that feature similarity carries most of the geometric signal while the attention cue mainly sharpens the reward.
Weighting the two equally ($5{:}5$) attains the best PF-ass (3.143) but degrades MET3R to 0.206, as the attention mass is diffuse and dilutes the geometric evidence.
We adopt $3{:}7$, which attains the best MET3R, QAlign, and DeQA at a marginal cost in PF-ass.

\begin{table}[t]
\centering
\small
\setlength{\tabcolsep}{5pt}
\begin{tabular}{l cccc}
\toprule
$w_s{:}w_t$ & QAlign$\uparrow$ & DeQA$\uparrow$ & MET3R$\downarrow$ & PF-ass$\uparrow$ \\
\midrule
0:1 & \underline{3.344} & \underline{3.993} & 0.209 & \underline{3.124} \\
3:7 & \textbf{3.381} & \textbf{4.012} & 0.184 & \textbf{3.133} \\
5:5 & 3.258 & 3.802 & \underline{0.176} & 2.999 \\
7:3 & 3.239 & 3.757 & \textbf{0.174} & 2.988 \\
\bottomrule
\end{tabular}
\caption{Ablation of the source and target weights $w_s{:}w_t$ in the geometry-consistency reward.
The two source-heavy settings reach a lower MET3R by under-editing, at a clear cost in composition and quality.}
\label{tab:abl_srctgt}
\end{table}

\noindent\textbf{Source-Target Reward Balance (Stage~II).}
Table~\ref{tab:abl_srctgt} varies the weights $w_s{:}w_t$ that balance source preservation against target alignment in the geometry reward.
Removing source preservation entirely ($0{:}1$) degrades MET3R to 0.209, confirming that the source term is what anchors the output to the input scene.
Conversely, raising the source weight to $5{:}5$ or $7{:}3$ does reach a lower MET3R (0.176 and 0.174), but at a sharp cost in composition and quality, e.g., PF-ass drops from 3.133 to 2.999 and 2.988.
These configurations improve MET3R by suppressing the edit rather than by better preserving geometry, which is precisely the under-editing bias of the metric noted above, and therefore do not represent a genuinely better trade-off.
We adopt $3{:}7$, which attains the best QAlign, DeQA, and PF-ass while keeping MET3R at 0.184.

In summary, both Stage~I design choices converge on the same conclusion, i.e., our configuration attains the best MET3R while remaining competitive on quality and composition, which is what this stage is meant to deliver, as its sole purpose is to instill geometric consistency into the representations.
In Stage~II, where the hybrid reward must balance geometry against instruction following and visual quality, our configuration instead attains the best overall trade-off across the four metrics, i.e., the best MET3R for the attention-feature balance, and the best QAlign, DeQA, and PF-ass for the source-target balance, in which the two source-heavy settings reach a lower MET3R only by suppressing the compositional change that the task requires.
This reflects the central design principle of GeoComposer, i.e., to prioritize geometric consistency without sacrificing the edit itself.

\end{document}